\pdfoutput=1
\documentclass[11pt,a4paper]{article}
\usepackage[hyperref]{emnlp2020}
\usepackage{times}
\usepackage{latexsym}

\usepackage{microtype}

\usepackage{color}
\usepackage{xspace}
\usepackage{booktabs}
\usepackage{graphicx}
\usepackage{amsmath}
\usepackage{amsfonts}
\usepackage{amssymb}
\usepackage{array}
\usepackage{multirow}
\usepackage{soul}
\usepackage[caption=false]{subfig}

\usepackage{url}

\newcommand{\isa}{\texttt{is-a}\xspace}
\newcommand{\bless}{\textsc{Bless}}
\newcommand{\wbless}{\textsc{WBless}}
\newcommand{\bibless}{\textsc{BiBless}}
\newcommand{\hyperlex}{\textsc{Hyperlex}}
\newcommand{\shwartz}{\textsc{Shwartz}}
\newcommand{\leds}{\textsc{Leds}}
\newcommand{\eval}{\textsc{Eval}}
\newcommand{\paratitle}[1]{\noindent \textbf{#1}}

\newcommand{\ie}{\emph{i.e.,}\xspace}
\newcommand{\eg}{\emph{e.g.,}\xspace}

\newcommand{\aka}{\emph{a.k.a.}\xspace}
\newcommand{\wrt}{\emph{w.r.t.}\xspace}

\newcommand{\tabincell}[2]{\begin{tabular}{@{}#1@{}}#2\end{tabular}}

\newcommand{\RN}[1]{%
  \textup{\uppercase\expandafter{\romannumeral#1}}%
}

\aclfinalcopy % Uncomment this line for the final submission
\title{When Hearst Is not Enough: Improving Hypernymy Detection \\ from Corpus with Distributional Models}

\author{Changlong Yu\textsuperscript{$1$}\thanks{~~~Work done when C. Yu, J. Han and P. Wang were with Tencent AI Lab.}\quad Jialong Han\textsuperscript{$2$}\footnotemark[1] \quad Peifeng Wang\textsuperscript{$3$}\footnotemark[1] \quad
Yangqiu Song\textsuperscript{$1$} \quad \\
\textbf{Hongming Zhang}\textsuperscript{$1$}\quad \textbf{Wilfred Ng}\textsuperscript{$1$}\quad \textbf{Shuming Shi}\textsuperscript{$4$} \\
\textsuperscript{$1$}HKUST \quad \textsuperscript{$2$}Amazon \quad \textsuperscript{3}University of Southern California \quad \textsuperscript{$4$}Tencent AI Lab \\
    \{cyuaq, yqsong, hzhangal, wilfred\}@cse.ust.hk \\ 
    jialonghan@gmail.com, \
    peifengw@usc.edu, \
    shumingshi@tencent.com \\
}

\date{}

\begin{document}
\maketitle
\begin{abstract}
We address hypernymy detection, \ie whether an \isa relationship exists between words $(x,y)$, with the help of large textual corpora.
Most conventional approaches to this task have been categorized to be either \emph{pattern-based} or \emph{distributional}.
Recent studies suggest that \emph{pattern-based} ones are superior, if large-scale Hearst pairs are extracted and fed, with the sparsity of unseen $(x,y)$ pairs relieved.
However, they become invalid in some specific sparsity cases, where $x$ or $y$ is not involved in any pattern.
For the first time, this paper quantifies the non-negligible existence of those specific cases.
We also demonstrate that distributional methods are ideal to make up for pattern-based ones in such cases.
We devise a complementary framework, under which a pattern-based and a distributional model collaborate seamlessly in cases which they each prefer. 
On several benchmark datasets, our framework achieves competitive improvements and the case study shows its better interpretability.

%our framework demonstrates improvements that are both competitive and explainable\textcolor{red}{[Consider replace this word]}.
\end{abstract}

% (1) hongming's citation 
% (2) bert experimental results
% (3) notice some narratives: explainable etc 
% (4) reorganize the introduction part 
% (5) 

\section{Introduction}\label{sec:intro}

A taxonomy is a semantic hierarchy of words or concepts organized \wrt their \emph{hypernymy} (\aka \isa) relationships.
Being a well-structured resource of lexical knowledge, taxonomies are vital to various tasks such as question answering~\cite{C18-1042}, textual entailment~\cite{dagan2013recognizing,bowman2015large,yu2020enriching}, and text generation~\cite{biran2013classifying}.
When automatically building taxonomies from scratch or populating manually crafted ones, the \emph{hypernymy detection} task plays a central role.
For a pair of queried words $(x_q,y_q)$, hypernymy detection requires inferring the existence of a hyponym-hypernym relationship between $x_q$ and $y_q$.
Due to the good coverage and availability, free-text corpora are widely used to facilitate hypernymy detection, resulting in two lines of approaches: \emph{pattern-based} and \emph{distributional}.

\begin{figure}[t]
  \centering
  \includegraphics[width=1.0\columnwidth]{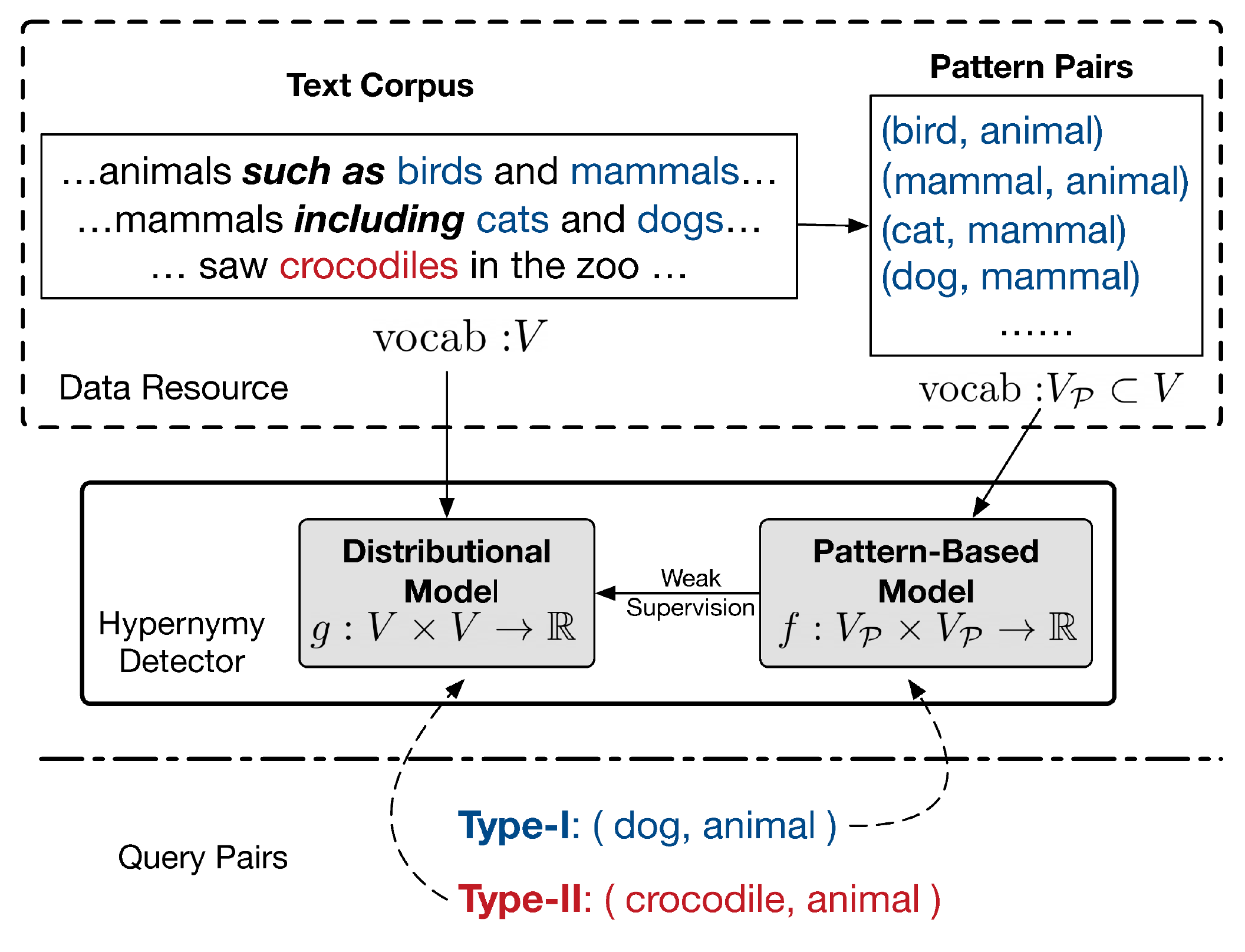}
  \caption{The overall framework of complementary methods for hypernymy detection from corpus. Different sparsity types of queried pairs are handled with pattern-based and distributional models respectively. }\label{fig:emnlp_framework}
\end{figure}

Pattern-based approaches employ pattern pairs $(x,y)$ extracted via \emph{Hearst-like patterns}~\cite{hearst1992automatic}, \eg ``$y$ \emph{such as}
$x$'' and ``$x$ \emph{and other} $y$''. 
An example of extracted pattern pairs from corpus are shown in Figure~\ref{fig:emnlp_framework}. 
Despite their high precision, the extracted pairs suffer from sparsity which comes in two folds \ie \textbf{Type-\RN{1}}: $x_q$ and $y_q$ separately appear in some extracted pairs, but the pair $(x_q,y_q)$ is absent \eg (dog, animal); or
\textbf{Type-\RN{2}}: either $x_q$ or $y_q$ is not involved in any extracted pair \eg (crocodile, animal). 

Although matrix factorization~\cite{P18-2057} or embedding techniques~\cite{vendrov2016order,nickel2017poincare,le2019inferring} are widely adopted to implement pattern-based approaches, they only relieve the \textbf{Type-\RN{1}} sparsity and cannot generalize to unseen words appearing in the \textbf{Type-\RN{2}} pairs.
On the other hand, distributional ones follow, or are inspired by, the \emph{Distributional Inclusion Hypothesis} (DIH; \citealt{geffet2005distributional}), \ie the set of the hyponym's contexts should be roughly contained by the hypernym's.
Although applicable to any word in a corpus, they are suggested to be inferior to pattern-based ones fed with sufficient extracted pairs~\cite{P18-2057,le2019inferring}.

Since pattern-based methods have unresolved sparsity issues, while distributional ones are more broadly applicable but globally inferior, neither of them can dominate the other in every aspect.
In this light, we are interested in two questions:
\begin{itemize}
\vspace{-0.5em}
\item Is the \textbf{Type-\RN{2}} sparsity severe in practice?
\vspace{-0.5em}
\item If so, how to complement pattern-based approaches with distributional ones where the former is invalid?
\end{itemize}
\vspace{-0.5em}

To answer the first question, we conduct analyses involving estimations on real-world corpora as well as statistics of common hypernymy detection datasets.
Results from both resources indicate that the likelihood of encountering the \textbf{Type-\RN{2}} sparsity in practice could even reach up to more than 50\%, which is thus non-negligible.

For the second question, we present ComHyper, a complementary framework~(Sec.~\ref{sec:4.1framework}) which takes advantage of both pattern-based models' superior performance on \textbf{Type-\RN{1}} cases and the broad coverage of distributional models on \textbf{Type-\RN{2}} ones.
%(\textcolor{red}{Old}: precise lexical knowledge from pattern-based models as well as broad coverage properties of distributional models.) 
Specifically, to deal with \textbf{Type-\RN{2}} sparsity, instead of directly using unsupervised distributional models, ComHyper uses a training stage~(Sec.~\ref{sec:4.3training}) to sample from output space of a pattern-based model to train another supervised distribution model implemented by different context encoders~(Sec.~\ref{sec:4.2encoder}).
%(\textcolor{red}{Old}: we bridge two models together by training distributional models, with supervision signals coming from pattern-based models in order to better cope with the \textbf{Type-\RN{2}} sparsity.)
%Here pattern-based models exactly provide precise hypernym prediction but are only capable of solving \textbf{Type-\RN{1}} sparsity.
%This design is inspired by \citet{shwartz2017hypernyms}, \ie supervised distributional models are superior to unsupervised ones, but require stable training signals.
%It also ensures that the output pairwise scores of the two models are comparable.
In the inference stage, ComHyper uses the two models to separately handle the type of sparsity they are good at, as illustrated in Figure~\ref{fig:emnlp_framework}. 
In this manner, ComHyper relies on the partial use of pattern-based models on \textbf{Type-\RN{1}} sparsity to secure performance no lower than distributional ones, and further attempts to lift the performance by fixing the former's blind spots (\textbf{Type-\RN{2}} sparsity) with the latter.
%enriches the distributional models by the knowledge from pattern-based ones and further works better on relieving the \textbf{Type-\RN{2}} sparsity.
On several benchmarks and evaluation settings, the distributional model in ComHyper proves effective on its targeted cases, making our complementary approach outperform a competitive class of pattern-based baselines~\cite{P18-2057}. 
Further analysis also suggests that ComHyper is robust when facing different mixtures of \textbf{Type-\RN{1}} and \textbf{-\RN{2}} sparsity.
%the two models affect and make up for each other, which motivates us to acquire lexical knowledge from both perspectives in a complementary but not isolated way. 

%In contrast to previous approaches that combine pattern-based and distributional models in the \textit{retrofitting} way~(\eg \citet{N18-1103, kamath-etal-2019-specializing}), our complementary approach achieves comparable performance efficiently without adjusting all the distributional vectors. 
%This also results in flexible continual learning, which distributional vectors need not be retrained when we need to add more pattern pairs. 
%Moreover only extracted pattern pairs are required in our framework rather than more linguistic constraints \eg synonym and antonym in \citet{N18-1103}.
%Thus our proposed approach provides a light-weight and flexible framework that integrates both types of models seamlessly and efficiently. 

%Moreover, if used in combination, a distributional model's scores on pairs with type-two sparsity should be comparable with the other pairs' scores from the pattern-based one.
%This is crucial in making the scores consistent and meaningful, if downstream systems involve them for calculation or ranking.

%The OOV model also needs some consideration.
%Previous studies~\cite{vulic2018post} suggest using general word embedding as input.
%We hypothesize that using context as input may facilitate the encoder to capture more distribution-based features than using word embeddings.

Our contributions are summarized as : 
\textbf{1)} We confirm that a specific type of sparsity issue of current pattern-based approaches is non-negligible.
\textbf{2)} We propose a framework of complementing pattern-based approaches with distributional models where the former is invalid.
%\item We propose depending DIH-based supervised embeddings on contexts rather than general word embeddings.
\textbf{3)} We systematically conduct comparisons on several common datasets, validating the superiority of our framework.

\section{Related Work}\label{sec:relatedWork}

%This section reviews conventional approaches on the hypernymy detection task, including pattern-based, distributional, and hybrid ones.
%We also discuss related studies that may inspire but are not yet applied in the task.

%Research on lexical entailment and hypernymy detection has attracted abundant attention in natural language processing community over decades. The ability of detecting and modeling hyponymy-hypernymy relations is critical for many related downstream applications such as question answering, language inference and text generation \cite{vulic2017hyperlex}. 

\paratitle{Pattern-Based Approaches.}
Taxonomies from experts (\eg WordNet~\cite{miller1995wordnet}) have proved effective in various reasoning applications~\cite{song2011short,DBLP:conf/acl/ZhangZS20}.
Meanwhile, Hearst patterns~\cite{hearst1992automatic} make large corpora a good resource of explicit \isa pairs, resulting in automatically built hypernymy knowledge bases~\cite{wu2012probase,seitner2016large} of large scales.
The coverage of both words and hypernymy pairs in those resources are far from complete.
%They cannot be directly applied for hypernymy detection via simple look-ups.  

To infer unknown hypernymies between known words, \eg implicit \isa pairs in transitive closures, pattern-based models are proposed. 
\citet{P18-2057} and \citet{le2019inferring} show that, on a broad range of benchmarks, simple matrix decomposition or embeddings on pattern-based word co-occurrence statistics provide robust performance.
On Probase~\cite{wu2012probase} - a Hearst-pattern-based taxonomy, \citet{yu2015learning} use embeddings to address the same sparsity problem.
Some methods~\cite{vendrov2016order,athiwilson_doe_2018,nickel2017poincare,nickel2018learning,pmlr-v80-ganea18a} embed WordNet in low-dimensional space.
% Although working on WordNet, we also categorize them as pattern-based because they share the same spirit of inferring unknown \isa pairs from known ones.
%Order embedding~\cite{vendrov2016order} embeds hypernyms and hyponyms so that the former's coordinates are smaller than the latter's in each dimension.
%\citet{athiwilson_doe_2018} extends order embedding by replacing word vectors with Gaussian word embeddings~\cite{vilnis2015word}, providing probabilistic interpretations for the prediction scores.
%Furthermore, hyperbolic embeddings adopt hyperbolic spaces to preserve distance and similarity within hierarchical graphs~\cite{nickel2017poincare,nickel2018learning,pmlr-v80-ganea18a}. 
Depending on vectors of words learnt from known \isa pairs, the above pattern-based methods cannot induce more hypernymy pairs whose words do not appear in any pattern.

%However these supervised embedding methods can only model hierarchical structure information and have weak generalization ability for unseen words or phrases.

%Traditional studies either advocate the coverage of DIH, or suggest~\cite{P18-2057} that pattern-based co-occurrence with matrix factorization is robust and can outperform distribution-based ones on several datasets.

%The starting point of solving lexical entailment tasks is manually predefined pattern rules to detect and bootstrap isA relations from large scale corpus \cite{hearst1992automatic}. For example patterns like "X such as Y" could be a strong signal to indicate the hypernymy relation between X and Y.  Such lexico-syntactic pattern-based methods are also well applied in word similarity tasks \cite{schwartz2015symmetric} and show great representation ability for lexical semantics \cite{schwartz2016symmetric}. However co-occurrence words are extremely sparse and those patterns could not generalize to unseen words. 

\paratitle{Distributional Approaches.}
Distributional models are inspired by DIH~\cite{geffet2005distributional}.
They work on only word contexts rather than extracted pairs, thus are applicable to any word in a corpus.
%researchers turn to use distributional representations for words, which are mostly based on \textsl{Distributional Inclusion Hypothesis}\cite{geffet2005distributional}. The intuition is that hyponymys are expected to occur in a subset of the contexts of their hypernymys.
Early unsupervised models typically propose asymmetric similarity metrics over manual word feature vectors for entailment~\cite{weeds2004characterising,clarke2009context,santus2014chasing}.
In \citet{chang2018distributional} and \citet{NguyenEtAl:17b}, the authors inject DIH into unsupervised embedding models to yield latent feature vectors with hypernymy information.
Those feature vectors, manual or latent, may serve in unsupervised asymmetric metrics or to train supervised hypernymy classifiers.
\citet{shwartz2017hypernyms} explore combinations of manual features and (un)supervised predictors, and suggest that unsupervised metrics are more robust \wrt the distribution change of training instances.
%Also using word vectors as features and designing complex neural network for supervised setting could achieve good performance
Projection learning~\cite{fu-etal-2014-learning,ustalov-etal-2017-negative,wang-he-2020-birre} has been used for supervised hypernymy detection. 

\paratitle{Other Improved Methods.}
Due to weak generalization ability of Hearst patterns, \citet{anh2016learning} and \citet{shwartz-goldberg-dagan:2016:P16-1} relieve the constraints from strict Hearst patterns to co-occurring contexts or lexico-syntactic paths between two words. 
They encode the co-occurring contexts or paths using word vectors to train hypernymy embeddings or classifiers.
Although leading to better recall than Hearst patterns~\cite{washio-kato-2018-filling}, they limit the trained embeddings or models from generalizing to every word in a corpus.
Nevertheless they have no ability to cope with the \textbf{Type-\RN{2}} sparsity, which is the main focus of our work.

Another line of retrofitting methods~\cite{vulic2018post,N18-1103},  \ie adjusting distributional vectors to satisfy external linguistic constraints, has been applied to hypernymy detection.
However, they strictly require more additional resources \eg synonym and antonym to achieve better performance~\cite{kamath-etal-2019-specializing}.
%In summary, previous hybrid methods only improve cases that pattern-based ones can address, while incurring additional restrictions.
%Finally, there are approaches that potentially generalize to all words in a corpus, but they are either evaluated only on limited datasets~\cite{P18-2101}, or not even on the hypernymy detection task~\cite{vulic2018post}. 
To the best of our knowledge, we are the first to propose complementing the two lines of approaches to cover every word in a simple yet efficient way, with extensive analysis of the framework's potential and evaluation of performances.
%of the practical performance.

%We do not improve type-1 patterns.
%They are hybrid rather than complementary.
%For the first question, to the best of our knowledge, few previous effort is spent along this line for hypernymy detection.
%In this paper, we adopt a hybrid approach of the two approaches at the instance level.
%Specifically, we use pattern-based approaches for word pairs involved or inferred in observed patterns, and use distribution-based methods for OOV pairs.
\section{Preliminaries}\label{sec:preliminaries}

We formally define the aforementioned two types of sparsity, 
and provide some statistical insights about their impacts on pattern-based methods.

\subsection{Notations and Definitions}

Let $V$ be the vocabulary of a corpus $\mathcal{C}$.
By applying Hearst patterns on $\mathcal{C}$, a set of \emph{extracted pairs} $\mathcal{P}\subseteq V\times V$, \ie \isa relationships $\{(x,y)\}$ ($x,y\in V$), is obtained.
As in Section~\ref{sec:relatedWork}, pattern-based approaches usually use $\mathcal{P}$ to perform matrix factorization or embedding learning.
Due to their nature, only words ``seen'' in $\mathcal{P}$, or $V_\mathcal{P}=\{x\mid(x,y)\in \mathcal{P}\vee (y,x)\in\mathcal{P}\}$, will have respective columns/rows or embeddings.
We refer to them by \emph{in-pattern} (or IP for short) words.
We refer to words without columns/rows or embeddings, \ie $V\setminus V_\mathcal{P}$, by \emph{out-of-pattern} (or OOP) words.

Suppose a pair of words $q=(x_q,y_q)$ is queried for potential hypernymy.
We say $q$ is an IP pair if both $x_p$ and $y_p$ are IP words, or an OOP pair if either of them is OOP. 
Due to the need of explicit columns/rows or embeddings for both $x_q$ and $y_q$, pattern-based approaches may only make inferences on IP pairs, but are infeasible on OOP ones.

\subsection{Observations and Motivation}

Given the infeasibility of pattern-based methods on OOP pairs, we are interested in what extent pattern-based methods are limited, \ie the rough likelihood of encountering OOP pairs in practice.
At first sight, Hearst patterns may have very sparse occurrences in a corpus.
Nevertheless, words with higher frequencies tend to be covered by Hearst patterns and be IP words.
Therefore, the possibility of encountering OOP pairs is not obvious to assess. 

\begin{figure}[t]
  \centering
  \includegraphics[width=.56\columnwidth]{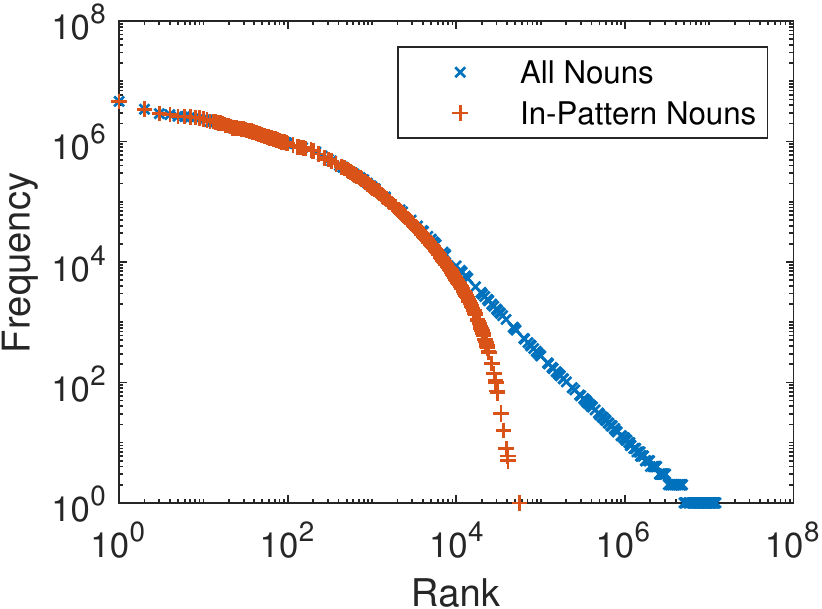}
  \caption{Corpus frequency of all nouns and IP nouns.}\label{fig:distribution}
\end{figure}

To shed light upon the OOP issue of pattern-based methods, we conduct an analysis on the corpora %\footnote{We replace the Gigaword corpus with ukWac, for the former is not complimentary.} 
and extracted pairs in \citet{P18-2057}. 
Considering that nouns tend to be queried more for potential hypernymy than, say, verbs, we only focus on nouns.
In Figure~\ref{fig:distribution}, we show the corpus frequency of all nouns and in-pattern nouns, and draw the following observations.

\textbf{1)} \textbf{$V_\mathcal{P}$ covers well the most frequent nouns in $V$.} For the top-10\textsuperscript{4} frequent nouns, the two lines of dots overlap well, indicating that common nouns are very likely to be involved in Hearst patterns.

\textbf{2)} \textbf{Due to the limited size of $V_\mathcal{P}$, it is unable to cover the tail of $V$.} With the frequency rank below 10\textsuperscript{4}, the two lines begin to separate. Comparing their intersections with the x-axis, it is understandable that a limited number of IP nouns cannot cover both frequent and tail nouns in a vocabulary, whose size is several orders of magnitudes larger.

\textbf{3)} \textbf{The likelihood of a noun being OOP is non-negligible.} The two lines enclose a triangular region, corresponding to the likelihood of a randomly drawn noun being OOP. According to our statistics, this region accounts for a non-negligible proportion of 19.9\% of the total area.

With the likelihood of OOP nouns at hand, we are ready to roughly estimate the likelihood of encountering OOP pairs in practice.
Suppose the two words in $q$ are nouns independently sampled from the corpus distribution.
Then the probability of $q$ being OOP, \ie infeasible for pattern-based methods, is $1-(1-0.199)^2=35.8\%$.
Even if $y_q$ tends to bias towards more common words, the optimistic estimation is still above 19.9\%.

Table~\ref{tab:detectionDataset} lists the actual portions of OOP pairs in several commonly used datasets \wrt $\mathcal{P}$ in \citet{P18-2057}.
Note that neither the datasets nor $\mathcal{P}$ are created in favor of the other.
These actual rates may be above or below the estimated interval of 19.9\%-35.8\%, but are all at considerable levels.
Considering the above analyses, we confirm that OOP pairs are non-negligible in practice and give a positive answer to the first question in Section~\ref{sec:intro}.

\paratitle{Motivation of the Study.}
OOP pairs are problematic for pattern-based methods.
Despite their non-negligible existence, former pattern-based methods~\cite{P18-2057,le2019inferring} boldly classify them as non-hypernymy in prediction.
However, distributional methods are applicable as long as the two queried words have contexts.
Thus, they are ideal to complement pattern-based methods on the non-negligible minority of OOP pairs.
%The motivation and remainder of this study will be devoted to our second question, \ie how to complement the two lines of methods.

\begin{table}[t]\small
\setlength\tabcolsep{4.4pt}
\centering
\begin{tabular}{lcrr}
\toprule
\textbf{Dataset}&\textbf{OOP (Hyper/All)}&\textbf{Total}&\textbf{OOP Rate}\\
\midrule
\bless&44 / 1,829&14,542&12.58\%\\
\eval&694 / 3,903&13,450&29.02\%\\
\leds&105 / 209&2,770&7.55\%\\
\shwartz&7,209 / 35,266&52,577&67.07\%\\
\textsc{W(Bi)Bless}&0 / 46&1,668&2.76\%\\
%\bibless&46&1,668&2.76 \\
\hyperlex&n/a / 107&2,163&4.95\%\\
\bottomrule
\end{tabular}
\caption{Statistics of OOP pairs \wrt extracted pairs $\mathcal{P}$. OOP~(All) is the number of OOP pairs while OOP~(Hyper) is the number of OOP with true labels. 
\label{tab:detectionDataset}}  
\end{table}

\section{Our Approach}\label{sec:approach}

%In light of the above analyses, this section describes our complementary framework.

\subsection{Framework}\label{sec:4.1framework}

Our framework is illustrated in Figure~\ref{fig:emnlp_framework}.
It consists of a \emph{pattern-based model} and a \emph{distributional model} cooperating on the \emph{data resource} to answer an arbitrarily queried pair of words $q\in V\times V$.

\paratitle{Data Resource.}
To train a pattern-based model using prior solutions, our data resource includes extracted pairs $\mathcal{P}$ from some text corpus $\mathcal{C}$.
Unlike pattern-based approaches that depend solely on $\mathcal{P}$, our data resource also involves the corpus $\mathcal{C}$ for the sake of the distributional model.

\paratitle{Pattern-Based Model.} 
The pattern-based model works on the extracted pairs $\mathcal{P}$ to serve in two roles.
On the one hand, it is responsible for generalizing from statistics on $\mathcal{P}$ to score any in-pattern pair $q\in V_\mathcal{P}\times V_\mathcal{P}$ to reflect the plausibility of a hypernymy relationship.
To this end, it is sufficient to adopt matrix-factorization-based~\cite{P18-2057} or embedding models~\cite{le2019inferring}.
On the other hand, the pattern-based model also provides supervision signals via a \emph{sampler} for training the distributional model.
We will specify this role later.
Formally, we denote the pattern-based model by $f:V_\mathcal{P}\times V_\mathcal{P}\rightarrow \mathbb{R}$.
%If we view WordNet as ideal patterns, LEAR is also a pattern-based model. Essentially pattern-based.
%\begin{figure}[t]
%  \centering
%  \includegraphics[width=0.88\columnwidth]{fig/framework.pdf}
%  \caption{Our complementary framework.}\label{fig:framework}
%\end{figure}

\paratitle{Distributional Model.}
Different from the pattern-based model defined on IP pairs $V_\mathcal{P}\times V_\mathcal{P}$, the distributional model has a form of 
$g:V\times V\rightarrow \mathbb{R}$, \ie it should be capable of predicting on any word pair in $V\times V$. 
This invalidates the model's dependency on extracted pairs involving $x_q$ or $y_q$.
The separate contexts of $x_q$ and $y_q$ in corpus $\mathcal{C}$ turn out to serve as the basis and input of the distributional model, respectively.
Given the superior performance of pattern-based models on IP pairs~\cite{P18-2057}, the distributional model $g$ is only responsible to answer OOP pairs.

Various choices exist to implement the distributional model.
We may apply unsupervised metrics~\cite{weeds2004characterising,clarke2009context,santus2014chasing} on manual features extracted from contexts of $x_q$ and $y_q$, which are robust to the distribution change of training data~\cite{shwartz2017hypernyms}.
However, the scores of those metrics are not necessarily in the same scale with those output by the pattern-based model $f$ for IP pairs.
Such inconsistency will harm downstream systems which involve the scores for ranking or calculation.

Given sufficient supervision signals from $f$ and the inherent noise of natural language, we implement the distributional model $g$ by a supervised neural-network-based approach.
%We denote $\mathbf{C}$ as the context encoder and abuse the notation of $g$ and further decompose it as \textcolor{red}{TODO: word2vec or context encoder.}
%Following DIH, we depend the distributional model on the contexts of a word to deal with OOP words.
%
%As specified in Figure~\ref{fig:framework}, for a word $x$, the context encoder works on a collection of contexts, and output $x$'s embedding.
%We apply a \emph{transforming function} $T(.)$ on $c$.
%After context embeddings are obtained, these transformed embeddings are fed to the aggregator $A$ to output an embedding $\mathbf{x}$ for the target word $x$.
%\textcolor{red}{Whether to use $A$ and $T$ depends on whether we have many $A$ and $T$ or complicated $A$ and $T$. Otherwise it is over-formulating.}
%\begin{equation}
%	\mathbf{C}(x)=A(\{T(c)\mid c\in C(x)\}).
%\end{equation}
Specifically, the network encodes the contexts of $x$ and $y$ in $\mathcal{C}$, \ie $\mathcal{C}(x)$ and $\mathcal{C}(y)$, to be $\mathbf{x}_h$ and $\mathbf{y}_H$, respectively, and makes predictions by a dot product, \ie
\[
g(x,y)=\langle\mathbf{x}_h,\mathbf{y}_H\rangle.
%g(x,y)=\frac{\langle\mathbf{x}_h,\mathbf{y}_H\rangle}{||\mathbf{x}_h||\cdot ||\mathbf{y}_H||}.
\]
Note that hypernymy is essentially asymmetric, so we distinguish $\mathbf{x}_h$ and $\mathbf{y}_H$ by the subscripts to reflect the asymmetry.
In practice, we adopt networks with separate parameters for $\mathcal{C}(x)$ and $\mathcal{C}(y)$, which is detailed in the next section.

%\[
%\min_{\mathbf{C},g} \ \mathbb{E}_{(x,y)\in \textcolor{red}{\mathcal{P}?}}\ %\Big(g(\mathbf{C}(x),\mathbf{C}(y))-f(x,y)\Big)^2
%\]
%Or Post-Specification-like:
%\[
%\min_\mathbf{C}\ \mathbb{E}_{x\in V_\mathcal{P}}\ ||\mathbf{C}(x)-LEAR(x)||_2
%\]

%\paratitle{The Post-Specification Paper.} Let $\mathbf{C}(x)$ be the pre-trained word vector, \eg word2vec, of $x$, and make it fixed during training. Use post-specification-like supervision.

\subsection{Encoding Queried Words}\label{sec:4.2encoder}

To implement the distributional model, we encode $\mathcal{C}(x)$ and $\mathcal{C}(y)$ into hypernymy-specific representations $\mathbf{x}_h$ and $\mathbf{y}_H$, respectively.
There are various off-the-shelf models to encode sentential contexts.
We take the following four approaches. 

\paratitle{Transformed Word Vector.}
Instead of working directly on the original contexts $\mathcal{C}(x)$ and $\mathcal{C}(y)$, this approach takes as input the pre-trained word vectors~\cite{mikolov2013distributed,pennington2014glove} $\mathbf{x}$ and $\mathbf{y}$ of $x$ and $y$, and apply two Multi-Layer Perceptrons (MLPs), respectively:
\[
\mathbf{x}_h=\mathbf{MLP}_h(\mathbf{x}),\quad \mathbf{y}_H=\mathbf{MLP}_H(\mathbf{y}).
\]
The intuition is that word vectors roughly depend on the contexts and encode the distributional semantics.
To make the MLPs generalize to $V$ rather than $V_\mathcal{P}$, the word vectors are fixed during training. 
Inspired by the \emph{post specialization} in ~\citet{vulic2018post}, it also takes a similar approach to generalize task-specific word vector transformations to unseen words, though their evaluation task is not hypernymy detection.

\paratitle{\textsc{NBoW} with \textsc{Mean}-Pooling.}
Given words $\{c_j\}_{j=1}^n$ in a context $c\in \mathcal{C}(x)$, the Neural Bag-of-Words (\textsc{NBoW} for short) encoder looks up and averages their pre-trained vectors $\mathbf{c}_j$ as $\mathbf{c}$, transforms $\mathbf{c}$ through a MLP, and averages the resulted vectors through a \textsc{Mean}-pooling layer as $\mathbf{x}_h$:
\[
\mathbf{x}_h=\frac{1}{|\mathcal{C}(x)|} \sum_{c\in \mathcal{C}(x)} \mathbf{MLP}_h(\mathbf{c}),\quad
\mathbf{c}=\frac{1}{n} \sum_{j=1}^{n} \mathbf{c}_j.
\]
%\begin{align}
%\mathbf{v}_C=\frac{1}{n} \sum_{i=1}^{n} \mathbf{c}_i\text{.}
%\end{align}
To obtain $\mathbf{y}_H$, a similar network is applied, though the two MLPs do not share parameters to reflect the asymmetry of hypernymy.
We fix the embeddings of context word vectors during training because satisfactory performance is observed.
Due to its simplicity, \textsc{NBoW} is efficient to train.
However, it ignores the order of context words and may not well reserve semantics.

\paratitle{\textsc{context2vec} with \textsc{Mean}-Pooling.}
To study the impacts of positional information within the context, we also attempt to substitute the \textsc{NBoW} with the \textsc{context2vec} encoder~\cite{melamud2016context2vec}.
In \textsc{context2vec}, two \textsc{LSTM}s are used to encode the left and right contexts $\overrightarrow{c}$ and $\overleftarrow{c}$ of an occurrence of $x$, respectively.
The two output vectors are concatenated as the final context representation $\mathbf{c}$ for the same transformation and averaging as for \textsc{NBoW}.
Formally,
\[
\mathbf{c}=\big[\ \overrightarrow{\mathbf{LSTM}}(\overrightarrow{c});\ \overleftarrow{\mathbf{LSTM}}(\overleftarrow{c})\ \big].
\]
Note that the encoder for $y$ still has separate parameters from those of $x$.
%We implement and compare it with \textsc{NBoW} as a counterpart.

\paratitle{Hierarchical Attention Networks.} 
\textsc{NBoW} and \textsc{context2vec} with \textsc{Mean}-Pooling both aggregate every context word's information into $\mathbf{x}_h$ and $\mathbf{y}_H$.
Given several long contexts and the fixed output dimension, it is
vital for encoders to capture the most useful information.
Inspired by \citet{yang2016hierarchical}, we incorporate attention on different words and contexts.
We use a feed-forward network to estimate the importance, and combine the information, of each context word to obtain $\mathbf{c}$:
\begin{align*}
\alpha_j=\text{softmax}\Big(\mathbf{w}_a^\top \tanh(\mathbf{W}_a \mathbf{c}_j)\Big)\text{,}\ \mathbf{c}=\sum_{j=1}^{n} \alpha_j \mathbf{c}_j\text{.}
\end{align*}
Then, another similar network is applied to all $\mathbf{c}^{(i)}\in \mathcal{C}(x)$ to obtain the representation of $\mathbf{x}_h$:
\begin{align*}
\beta_i=\text{softmax}\Big(\mathbf{w}_b^\top \tanh(\mathbf{W}_b \mathbf{c}^{(i)})\Big)\text{,}\ \mathbf{x}_h=\sum_{i=1}^{|\mathcal{C}(x)|} \beta_i \mathbf{c}^{(i)}\text{.}
\end{align*}
For word $y$, the encoder is similar but still has separate parameters from those of $x$.

\subsection{Training the Distributional Model}\label{sec:4.3training}

We train the distributional model $g$'s parameters $\mathbf{\Phi}$ with supervision signals from the pattern-based model $f$.
To make output scores of $f$ and $g$ comparable, we adopt the square error between the two scores as the loss on a pair $(x,y)$, \ie
\[
l(x,y;\mathbf{\Phi})= \Big(g(x,y;\mathbf{\Phi})-f(x,y)\Big)^2.
\]

Compared with the potentially large size of the output space, a set of random samples from it suffices to train the parameters $\mathbf{\Phi}$.
For each IP word $x\in V_\mathcal{P}$, we uniformly sample $k$ entries from $\Delta_x$, the column and row involving $x$ in the output space $V_\mathcal{P}\times V_\mathcal{P}$:
%$we need both positive triplets and negative ones.
%All extracted pairs $\mathcal{P}$ from the knowledge graph naturally serve as positive pairs.
%We denote the set of positive triplets from $\mathcal{K}$ as $\Delta$ and the set of negative triplets as $\Delta^{'}$. Since the knowledge graph only contains positive triplets, 
%To make up for the absence of negative pairs, for each $(x,y)\in\mathcal{P}$, we randomly corrupt the hyponymy or hypernymy (but not both) by another word in $V_\mathcal{P}$, and denote the corresponding negative pairs by $\Delta'_{(x,y)}$. Formally,
%\[
%\Delta'_{(x,y)}=\{(x',y)\mid x'\in V_\mathcal{P}\}\ \cup\ \{(x,y')\mid y'\in V_\mathcal{P}\}.
%\]
\[
\Delta_x=\{(x,y)\mid y\in V_\mathcal{P}\}\ \cup\ \{(y,x)\mid y\in V_\mathcal{P}\}.
\]
%Let $P_{(x,y)}$ be a uniform distribution over $\Delta'_{(x,y)}$.
The sample for $x$ is done on $P_x$, a uniform distribution over $\Delta_x$.
Finally, our objective is
%\begin{eqnarray}\label{eq:object_function}
%\min\sum\limits_{(x,y)\in\mathcal{P}}
%\mathcal{L}(x,y;\mathbf{C}, g),\\
\[
\min\sum\limits_{x\in V_\mathcal{P}}
\mathcal{L}(x;\mathbf{\Phi}),
\]
%where $\mathcal{L}(x,y;\mathbf{C}, g)=$
%\begin{align*}
%l(x,y;\mathbf{C}, %g)+\sum_{i=1}^{k}\mathbb{E}_{(x^{(i)},y^{(i)})\sim P_{(x,y)}}l(x^{(i)},y^{(i)};\mathbf{C}, g)
%\end{align*}
where $\mathcal{L}(x;\mathbf{\Phi})$ is the expected loss related to $x$:
\begin{align*}
\mathcal{L}(x;\mathbf{\Phi})=\sum_{i=1}^{k}\mathbb{E}_{(x^{(i)},y^{(i)})\sim P_x} l(x^{(i)},y^{(i)};\mathbf{\Phi}).
\end{align*}

\section{Experimental Setup}

We adopt the widely-used comprehensive evaluation framework\footnote{\url{https://github.com/facebookresearch/hypernymysuite}} provided by \citet{P18-2057,le2019inferring}. To make experimental results comparable, we align the settings as much as possible.
%For clarity, this section briefly reviews our common corpora, tasks, datasets, evaluation protocols, and baselines, and explains the difference if any.

\subsection{Corpora and Evaluation}

\paratitle{Corpora.} We used the 431k \isa pairs (243k unique) released by~\citet{P18-2057}. 
We substitute the Gigaword corpus they used by uKWac~\cite{ferraresi2007building} because the former is not complimentary. This decision does not affect reproducing pattern-based approaches in \citet{P18-2057}.

\paratitle{Evaluation Tasks.} The three sub-tasks include \textbf{1)} ranked hypernym detection: given $(x_q, y_q)$ decide whether $y_q$ is a hypernym of $x_q$. Five datasets \ie   \bless~\cite{baroni2011we},
\eval~\cite{santus2015evalution},
\leds~\cite{baroni2012entailment}, \shwartz~\cite{shwartz-goldberg-dagan:2016:P16-1} and \wbless~\cite{weeds2014learning} are used. The positive predictions should be ranked higher over negative ones and \emph{Average Precision} (AP) is used for evaluation. \textbf{2)} hypernymy direction classification: determine which word in a pair has a broader meaning. Besides \bless~ and \wbless, we also use \bibless~\cite{kiela2015exploiting} and \textit{Accuracy}~(Acc.) is reported for binary classification.
\textbf{3)} graded entailment: predict scalar scores on  \hyperlex~\cite{vulic2017hyperlex}. Spearman's correlation $\rho$ between the labels and predicted scores is reported. 

The statistics of datasets are shown in Table~\ref{tab:detectionDataset}. 
The three tasks require algorithms to output scores unsupervisedly, which indicate the strength of hypernymy relationships.
Note no external training data is available in the evaluation. Only extracted Hearst pattern pairs may be used for supervision. 

\subsection{Compared Methods}

\paratitle{Pattern-Based Approaches.}
We reproduce four pattern-based methods
%\footnote{The original training/development/testing folds are not published so we had to generate ours. The reproduced results are comparable with those in \citet{P18-2057}.} in \citet{P18-2057}, 
\ie Count, PPMI, SVD-Count, and SVD-PPMI.
%Count simply returns the frequency of a pair in $\mathcal{P}$, while PPMI returns the log-frequency normalized by the respective frequency of both words, or zero if the score is negative.
%SVD-Count and SVD-PPMI scores are obtained by SVD decomposing and reconstructing the Count and PPMI matrices.
As in \citet{P18-2057}, SVD-PPMI is generally the most competitive.

\paratitle{Distributional Approaches.}
We compare with unsupervised distributional baselines in \citet{P18-2057}, \ie Cosine, Weeds Precision (WP), invCL, and SLQS.
%They are computed between context-based feature vectors of words.
%We follow the same manner to build the feature vectors and compute the scores as in \citet{P18-2057}.
%We refer readers to the paper for details of the metrics and features.
For supervised distributional baseline, we adopt the strongest model SDSN in~\citet{P18-2101} and take the probability scores of binary classifier as hypernymy predictions. All the 431k extracted pairs serve as true hypernymy pairs and false ones are generated by replacing one of the terms in true pairs with a random term.    
%Although we adopt supervised distributional models for complementing, our framework is overall unsupervised.
%and testing datasets are all unsupervised ones.
%Therefore, we do not compare with supervised methods~\cite{shwartz-goldberg-dagan:2016:P16-1,yu2015learning,anh2016learning}.

%To create semantic spaces for distributional models, we employ the large distributional space of \newcite{shwartz2017hypernyms} built from  a concatenation of two corpora: ukWac and WaCkypedia\_EN\cite{baroni2009wacky}. We adopt indirectional window-based contexts to construct co-occurrence matrix and them PPMI transformation is applied to it. The window size is set to 5 and words which appear less than 100 times are filtered.

\paratitle{Complementary Approaches.}
We adopt SVD-PPMI as the pattern-based model in our framework.
We pre-train 300-dimensional word embeddings with Skip-Gram~\cite{mikolov2013distributed} on our corpus for the use of the distributional model.
Specifically, we compare transformed word vector (\textsc{W2V}), \textsc{NBoW}/\textsc{context2vec} with \textsc{Mean}-Pooling (\textsc{NBoW}/\textsc{C2V}), and Hierarchical Attention Networks (\textsc{HAN})\footnote{Heavy contextualized encoders based on the pretrain-finetune framework did not yield considerable improvement and we focus on efficient traditional encoders which already outperform the baselines. Though we include the BERT encoders in our released code, we suggest to make tradeoffs when choosing encoders as discussed in~\citet{xia2020bert}.}.
The output dimension of our four encoders is set to 300.
The batch size is set to 128 and learning rate to 10\textsuperscript{-3}. 
We tuned the sampling size $k$ in $\{1, 3, 5 , 10, 100, 200, 400, 800\}$ on the validation set.
We did not tune other hyper-parameters since the default settings work well.
Our code is available at \url{https://github.com/HKUST-KnowComp/ComHyper}. 
%Specifically, the optimal $k$ is 400 for \textsc{W2V} and 1 for \textsc{NBow} and \textsc{C2V}. 

\begin{table}[t]\small
\setlength\tabcolsep{1pt}
\centering
\begin{tabular}{lcccccc}
\toprule
     & \multicolumn{4}{c}{{\bf Detection} (AP)}  & \multicolumn{1}{c}{{\bf Dir.}(Acc.)}& {\bf Graded}($\rho$)\\ \cmidrule(r){2-5} \cmidrule(lr){6-6}\cmidrule(l){7-7}
     &\bless&\eval&\leds&\shwartz&\bless&\hyperlex\\ \midrule

Cosine & .106 & .172 & .736 & .175 & .000 & -0.107\\
WP & .100 & .251 & .880 & .283 & .636 &0.147 \\
invCL  & .096 & .211 & \textbf{.887} & .220 & .636 & 0.062      \\
SLQS & .020 & .166 & .423 & .240 & .341 & -0.130 \\
\midrule
\textsc{W2V} & .292 & .255 & .712 & .453 & .767 & 0.313\\ 
\textsc{NBoW} & .124 & \textbf{.258} & .617 & .500 &  \textbf{.975}  & 0.264\\
\textsc{C2V} & .027 & .258 & .659 & .364 & .791 & \textbf{0.346}\\
\textsc{HAN} & \textbf{.346} & .250 & .602 & \textbf{.574} & \textbf{.975} & 0.309\\

\bottomrule

\end{tabular}
\caption{\label{tab:oov_res} Experimental results on OOP pairs.}  
\end{table}

\section{Experimental Results}\label{sec:experiments}

We aim to answer:
\textbf{1)} Are our distributional models supervised well by the pattern-based model?
\textbf{2)} Do they improve our complementary methods over the pattern-based ones?
\textbf{3)} Are complementary methods robust \wrt fewer extracted pairs?

\begin{table*}[t]\small
\centering
\setlength\tabcolsep{4pt}
\begin{tabular}{lcccccccccc}
\toprule
    & & \multicolumn{5}{c}{{\bf Detection} (AP)}  & \multicolumn{3}{c}{{\bf Direction} (Acc.)} & {\bf Graded} ($\rho$) \\  \cmidrule(r){3-7} \cmidrule(lr){8-10} \cmidrule(l){11-11}
     &  & {\bless} & \eval &  \leds &  {\shwartz} &  {\wbless} & \bless & \wbless & \bibless & \hyperlex   \\ \midrule
\multirow{4}{*}{Pattern}
& Count & .486 & .368 & .710 & .288 & .744 & .466 & .690 & .617 &  \textbf{.617} \\
& PPMI & .448 & .341 & .707 & .277 & .734 & .466 & .682 & .611 &  \underline{.603} \\
& SVD-Count & .651 & .434 & .812 & .369 & .904 & .936 & .842 & .801 &  .518 \\
%spmi~\cite{P18-2057}& .76 & .48 & .84 & .44 & .96 & .96 & .87 & .85 &.53\\
& SVD-PPMI & .764 & .463 & .831 & .409 & .959 & .959 & .871 & .847 & .517  \\
\midrule
Supervised & SDSN & .749 & .458 & .841 &  .432  & .958 & .959 &.874 & .851 & .588 \\ 
\midrule
\multirow{5}{*}{\begin{minipage}{0.6in}ComHyper (\textit{Ours})\end{minipage}}
& \textsc{W2V} & \ \textbf{.773} &  \underline{.474} & \textbf{.845} & .509 & .957 & .963 & .873 & .849 & .522 \\

& \textsc{NBoW} & .770 & \textbf{.474} & \underline{.844} & \underline{.510} & .958 & \underline{.970} & \underline{.875}  & \textbf{.853} & .523 \\ 

& \textsc{C2V} & .767 & .472 & .843 & .480 & \underline{.959} & .966 & .872  & .847 & .521 \\ 
%\textsc{SAN} & .769 & \textbf{.475} & .841 & \underline{.511} & \underline{.959} & .970 & \underline{.875}  & .853 & .525
%\\
& \textsc{HAN} & \underline{.772} & .473 & .843 & \textbf{.515} & \textbf{.959} & \textbf{.971} & \textbf{.875}  & \underline{.853} & .525 \\ \cmidrule(l){2-11}

& Oracle& .801 & .666 & .876 & .861 & .959 & .992 & n/a & n/a & n/a \\

\bottomrule

\end{tabular}
\caption{\label{tab:expresult} Experimental results on all queried pairs. Best ones are marked bold while second-best ones underlined.}  
\end{table*}

\subsection{Performance on OOP Pairs}

To ensure that our supervised distributional models are working effectively on OOP pairs, we evaluate on only OOP pairs under the aforementioned settings.
Because pattern-based approaches trivially give the lowest scores to OOP pairs, we only compare with distributional approaches.

Table~\ref{tab:oov_res} demonstrates the results.
Note that the 46 OOP pairs in \wbless\ and \bibless\ are all labeled false, causing undefined AP and perfect Acc.\ scores, so we omit the corresponding columns to save space.
Observing from Table~\ref{tab:oov_res}, except on \leds, our distributional models generally achieve higher scores than unsupervised approaches.
Especially, on the \bless\ dataset, Cosine even gets a zero Accuracy score because it is symmetric and cannot suggest the right direction.
The higher AP and Accuracy scores suggest that, supervised by the pattern-based model, our distributional models can generate better relative rankings within the scope of OOP pairs.

%Instead of human-crafted asymmetric hypernymy measurements, our proposed framework benefits from supervision signals harvested from Hearst-patterns. Results in Table \ref{oov_res} shows the superiority of context aggregation representation over large sparse distributional  space. Moreover the dimension of unsupervised sparse representation is up to million, which severely affects the efficiency of real-life prediction and downstream tasks. 

\subsection{Main Results and Case Study}

When facing both IP and OOP pairs, it is not enough to rank both types of pairs separately, since downstream systems usually require comparable scores or a unified ranking.
We evaluate on the entire datasets under the aforementioned settings.
We only compare with pattern-based methods and supervised distributional models because they generally outperform unsupervised ones.

Table~\ref{tab:expresult} provides the main results.
Best results are marked bold, and second-best ones are underlined.
To better interpret the results, we also provide ``Oracle'' scores, \ie the upper-bounds that complementary methods can achieve.
For the Detection task, Oracle scores are obtained by assigning OOP pairs having hypernymy relationships (See Table \ref{tab:detectionDataset}) the maximum score and other ones the minimum.
For \bless\ of Direction, the Oracle score is computed by assuming perfect predictions for OOP pairs.
The Oracle scores for \wbless/\bibless\ of the Direction task and \hyperlex\ of Graded Entailment are not straightforward to estimate, thus are omitted.
%In addition we provide the upper-bounds of our complementary approaches for exhaustive comparison. For five datasets of hypernymy detection tasks, pattern-based methods simply assign lowest score of all IP pairs to OOP pairs, which  

In Table~\ref{tab:expresult}, complementary methods lead to superior results on Detection and Direction tasks.
In eight out of nine columns, the best and second best scores are both achieved by complementary methods.
Especially, large improvements (up to 25.9\%) are observed on \shwartz\ with a higher OOP rate and thus a higher Oracle. 
%Moreover, the \textsc{C2V} context encoder generally outperforms \textsc{W2V} and \textsc{NBOW} due to its better capability of encoding sequential and positional information. 
In general, the \textsc{HAN} encoder achieves better performances.
By attending to the most informative contexts and words, the \textsc{HAN} encoder potentially captures distributional semantics that are relevant to hypernymy relationships between queried words.
Note that the relative performances between different context encoders are not necessarily consistent with those in Table~\ref{tab:oov_res}.
This is because the overall performance is not only sensitive to the relative ranking of OOP pairs, but also to their absolute scores.

In addition, with the same extracted $\mathcal{P}$ as supervision signals, our proposed methods show a great superiority over the supervised method~(SDSN in Table~\ref{tab:expresult}). 
Both SDSN and our complementary approaches could be regraded as \textit{combining pattern-based and distributional model}. 
The key difference is that complementary methods solve Type-I sparsity with a pattern-based model, which proved to be better than distributional ones on this case, while SDSN uses a distributional model (though supervised) uniformly on both cases.

\begin{figure*}[t]
  \centering
  \includegraphics[width=0.97\textwidth]{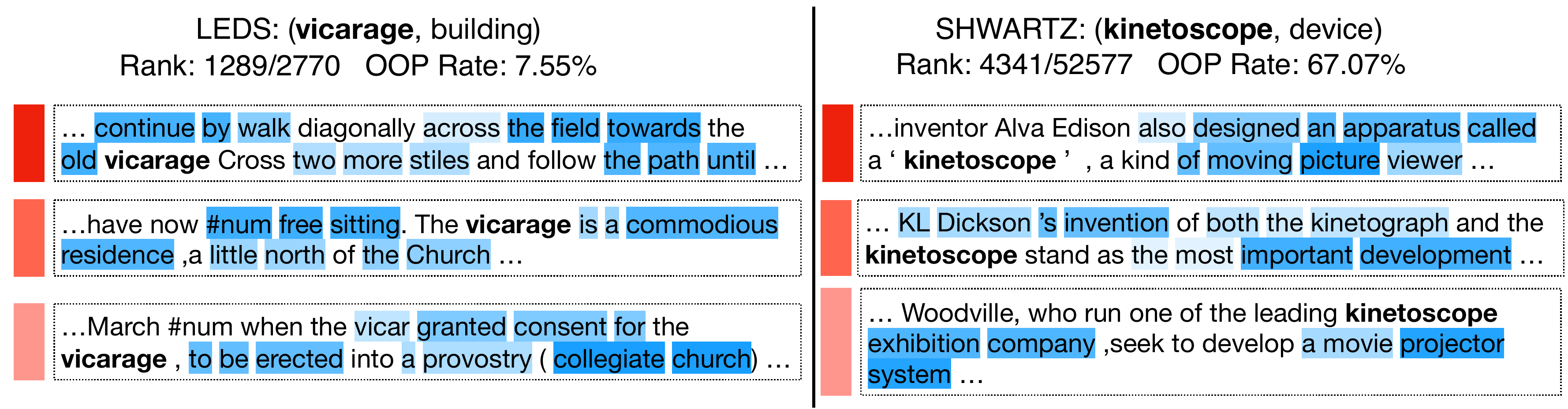}
  \caption{Case study of two queried pairs from two datasets, with OOP rates and actual ranks.}\label{fig:casestudy}
\end{figure*}

\begin{figure}
%\vspace{2mm}
\centering
\subfloat[Detection on \eval]{\label{fig:reduced_eval}
\includegraphics[scale=0.55]{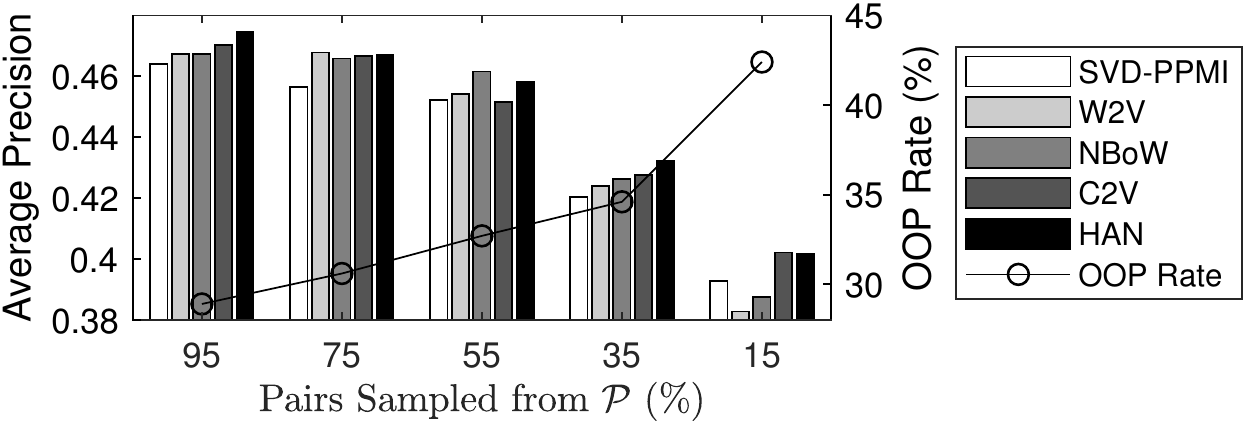}}
\
 \subfloat[Direction on \bless]{\label{fig:reduced_bless}
\includegraphics[scale=0.58]{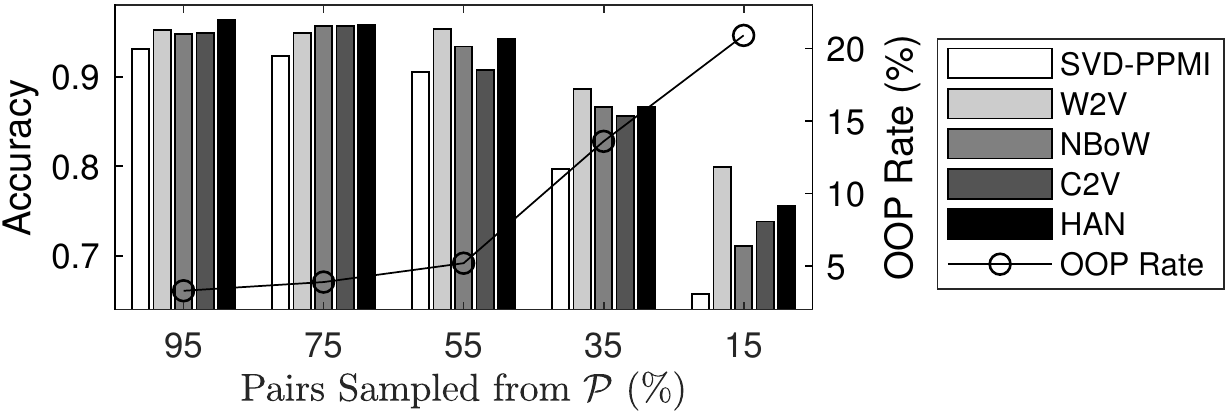}}
%\begin{minipage}[b]{0.19\columnwidth}
%\centering
%\subfloat[Legend]{
%\includegraphics[scale=0.45]{fig/legend.pdf}}
%\end{minipage}
\caption{Performance comparison across different amounts of reducing pairs on \eval\ and \bless.}
\label{fig:reducedPattern}
\vspace{-1em}
\end{figure}

\noindent \textbf{Case Study.} To explain the superiority of the \textsc{HAN} encoder, we exemplify with two true-hypernymy OOP pairs from two Detection datasets, respectively.
Here, the two hyponyms are both uncommon and OOP words.
Therefore, pattern-based models such as SVD-PPMI simply assign the pairs with minimum scores and rank them at the bottom.
But by examining their contexts in the textual corpus, the hypernymy relationships could have been inferred, and they could have been scored higher.

In Figure~\ref{fig:casestudy}, we show the two OOP pairs, as well as their rank according to \textsc{HAN} and the OOP rates of the corresponding datasets.
We also demonstrate the Top-3 contexts scored by \textsc{HAN} and visualize the context- and word-level attention weights.
We observe that \textsc{HAN} can attend to informative contexts and words that help capture the semantics of the OOP word.
For example, in \leds, \textbf{vicarage} is OOP.
\textsc{HAN} suggests three contexts that imply its meaning well.
By reading the context words and phrases highlighted by \textsc{HAN}, \eg \textit{commodious residence}, and \textit{collegiate church}, even people not knowing the word may guess it is a type of building.
With our \textsc{HAN}-based distributional model, the pair is successfully promoted to top 50\% in the ranking, well out of and above the bottom 7.55\% of OOP pairs.
Similar observations are drawn for the other pair, \ie (\textbf{kinetoscope}, device) with contexts \textit{moving picture viewer}, and \textit{movie projector system}.

We also observe that wrong predictions may be caused by extremely sparse contexts in the corpus such as \emph{famicom} in the dataset \shwartz. 

%found in the second pair \textit{(corkscrew, implement)} from \bless,  the meaning of rare word \textbf{corkscrew} with its context \textit{open bottle},\textit{knife}, \textit{stainless steel} can be captured by our context encoder and its rank gets improved \textbf{74\%}  compared with baseline model. The third pair \textit{(calamus,specie)} is sampled from \shwartz, which

%Our context encoder provides semantic complement for OOP words and automatically selects most relevant information for hypernymy detection from large corpus.  Furthermore our framework gives reasonable prediction scores for OOP pairs so that the overall rank of True cases would be increased while False ones still keep relatively low in terms of average precision evaluation.  It could be observed that the rank  of pair \textit{(vicarage, building)} makes a great leap from  \textbf{92.71\%} to  top \textbf{30\%}.   

%Based on the above quantitative and qualitative results, we confirm that, by addressing OOP pairs that pattern-based approaches fail to, our distributional models complement pattern-based models and improve the overall performance.

\subsection{Impacts of Reduced Pairs}

To analyze our complementary framework's robustness \wrt sparser extracted pairs $\mathcal{P}$, we randomly sample \{95\%,75\%,55\%,35\%,15\%\} of all 243k \isa pairs, and rerun SVD-PPMI, the best pattern-based approach and our complementary approaches.
In Figure~\ref{fig:reducedPattern}, we only illustrate the results on \leds\ for Detection and \bless\ for Direction.
Observations on the other datasets are similar, thus are omitted.
We have the following observations.
First, with fewer extracted pairs, the OOP rates increase quickly, and all models generally perform worse.
%resulting in another 5 sets of extracted pairs with increasing OOP rates. Then we train our complementary approaches for each of the resulting sets. The experimental results are displayed in Figure~\ref{fig:reducedPattern}, where we only report results on \leds, \shwartz and \bless datasets and omit the ones on other datasets due to the similar observations. 
This is not surprising since a sparser $\mathcal{P}$ leads to a less informative SVD-PPMI matrix and less supervision on distributional models.
%The only exception is in Figure~\ref{fig:reduced_shwartz}, where our complementary methods are even better when the OOP rate is higher than 70\%.
%We observe that, on this dataset, our distributional models are better than SVD-PPMI even on all IP pairs.
%When more IP pairs become OOP, they are forwarded to the distributional models, increasing the overall scores.
%with OOP lower than 70\%, there's still a considerable part of the testsets which are predicted by the pattern-based part of our framework, which is getting worse with less observed patterns. However, with OOP higher than 70\%, most of the testsets are predicted by the distributional part of our framework, which does not get worse as much as the pattern-based part does. 
%First of all, we could observe that with the OOP rate increasing, the performances of SVD-PPMI and our three models are majorly getting worse (except in Figure~\ref{fig:reduced_shwartz} with over 70\% OOP rate). This is not surprising since with less observed patterns the SVD-PPMI would certainly fail to approximate the desired scores for the unobserved patterns. Thus, the pattern-based part of our framework would be less reliable and then further affect the overall performance. 
Second, despite the increased OOP rates, our complementary methods consistently outperform SVD-PPMI
%Specially, in Figures~\ref{fig:reduced_shwartz} and~\ref{fig:reduced_bless}, 
and suffer less from increasing OOP rates especially on \bless.
%Moreover, under whichever the OOP rate, our three complementary approaches all outperform the pattern-based SVD-PPMI. From the Figure~\ref{fig:reduced_shwartz} and~\ref{fig:reduced_bless}, we could observe that our approaches suffer less from the increasing OOP rate than SVD-PPMI. This demonstrates that our proposed framework is more robust to the increasing OOP rates and proves their necessity in complementing the pattern-based approaches.
Finally, among the four context encoders, \textsc{HAN} performs better than the others when the sampled rate is higher than 75\%. However, with lower sampled rates, \textsc{W2V} is more robust than the others on \bless\ but fails to exceed \textsc{HAN} on \eval.

%\subsection{Error Analysis and Discussion}

%\draftycl{
%We identify some of the errors and limitations of our approach as well as directions for the future work.  
%}

%\textsc{W2V} is the most robust \wrt the increasing OOP rate.
%We attribute this to its smaller parameter space and less demand for training signals.
%while the performances of both NBoW and C2V are worse compared with W2V. This might due to the fact that the larger capacity neural models need more training signals, which is a problem also met by other supervised learning tasks. 

\section{Conclusion and Future Work}\label{sec:conclusion}

We propose complementing pattern-based and distributional methods for hypernymy detection.
As far as we know, this is the first work along this line.
We formally depict two types of sparsity that extracted pairs face, and indicate that pattern-based methods are invalid on the \textbf{Type-\RN{2}}, \ie out-of-pattern pairs.
By analyzing common corpora and datasets, we confirm that OOP pairs are non-negligible for the task.
To this end, we devise a complementary framework, where a pattern-based and distributional model handle IP and OOP pairs separately, while collaborating seamlessly to give unified scores.
%We propose alternatives of the distributional model's context encoder.
Oracle performance analysis shows that our framework has high potentials on several datasets.
Supervised by the pattern-based model, the distributional model shows robust capability of scoring OOP pairs and pushing the overall performance towards the oracle bounds.

In the future, we will extend the similar approach to multilingual~\cite{yu-etal-2020-hypernymy} or cross-lingual~\cite{upadhyay-etal-2018-robust} lexical entailment tasks. 
Moreover, one interesting direction is to use hyperbolic embeddings~\cite{le2019inferring,balazevic2019multi} for pattern-based models due to their inherent modeling ability of hierarchies.

\section*{Acknowledgements}

This paper was partially supported by the Early Career Scheme (ECS, No. 26206717), the General Research Fund (GRF, No. 16211520), and the Research Impact Fund (RIF, No. R6020-19) from the Research Grants Council (RGC) of Hong Kong.

\bibliography{emnlp2020}
\bibliographystyle{acl_natbib}

\end{document}